\documentclass[letterpaper, 10 pt, conference]{ieeeconf}

\usepackage{epsfig}
\usepackage{amssymb}
\usepackage{cite}
\usepackage{booktabs}
\usepackage{multirow}
\usepackage{mathtools}
\usepackage{bm}
\usepackage{algorithm}
\usepackage{comment}
\usepackage{color}
\usepackage{setspace}

\newcommand{\figlab}[1]{\label{fig:#1}}
\newcommand{\figref}[1]{Fig.~\ref{fig:#1}} 
\newcommand{\tablab}[1]{\label{tab:#1}}
\newcommand{\tabref}[1]{Table~\ref{tab:#1}} 
\newcommand{\forlab}[1]{\label{for:#1}}
\newcommand{\forref}[1]{Equation~(\ref{for:#1})} 

\newcommand{\etal}{\textit{et~al.}}
\definecolor{green}{rgb}{0.01, 0.5, 0.01}

\usepackage{amssymb}
\usepackage{pifont}
\newcommand{\cmark}{\ding{51}}%
\newcommand{\xmark}{\ding{55}}%

\usepackage{subcaption}
\usepackage{url}
\usepackage{threeparttable}

\usepackage{amsmath}
\usepackage[noend]{algorithmic}
\usepackage{array}

\begin{document}
\title{Shell-Type Soft Jig for Holding Objects during Disassembly}
\author{Takuya Kiyokawa$^{1}$, Ryunosuke Takebayashi$^{1}$, and Kensuke Harada$^{1,2}$, 
\thanks{$^{1}$Department of Systems Innovation, Graduate School of Engineering Science, The University of Osaka, 1-3 Machikaneyama, Toyonaka, Osaka, Japan.}%
\thanks{$^{2}$Industrial Cyber-physical Systems Research Center, The National Institute of Advanced Industrial Science and Technology (AIST), 2-3-26 Aomi, Koto-ku, Tokyo, Japan.}%
}

\maketitle

\vspace{-20pt}
\begin{abstract}
This study addresses a flexible holding tool for robotic disassembly. We propose a shell-type soft jig that securely and universally holds objects, mitigating the risk of component damage and adapting to diverse shapes while enabling \textit{soft fixation} that is robust to recognition, planning, and control errors.
The balloon-based holding mechanism ensures proper alignment and stable holding performance, thereby reducing the need for dedicated jig design, highly accurate perception, precise grasping, and finely tuned trajectory planning that are typically required with conventional fixtures.
Our experimental results demonstrate the practical feasibility of the proposed jig through performance comparisons with a vise and a jamming-gripper-inspired soft jig. 
Tests on ten different objects further showed representative successes and failures, clarifying the jig’s limitations and outlook.
\end{abstract}

\IEEEpeerreviewmaketitle

\section{Introduction}
As part of efforts toward realizing a sustainable society, robotic disassembly of various small electronic appliances has gained attention as a means of promoting automation in remanufacturing. Disassembly requires removing components from multiple directions and efficiently extracting valuable internal parts.

In robotic assembly and disassembly, high-precision operations are often supported by jigs that constrain target components in predetermined poses. Conventional rigid jigs, however, must be designed and fabricated for each specific geometry, limiting flexibility and scalability. To address diverse demands, there has been growing interest in general-purpose jigs that can adapt to multiple objects.

Several approaches have been investigated, such as vises with adjustable orientations~\cite{vise} and fixtures employing pin arrays with shape-memory mechanisms~\cite{pinarray3,pinarray,pinarray2}. While these methods offer versatility, they often allow the object’s pose to vary, requiring highly accurate perception and motion planning.
Flexible jigs using deformable materials have also been proposed~\cite{Brown2010,Kiyokawa2021jig,Sakuma2022}, but their fixation accuracy is typically lower than that of rigid jigs, making stable holding during disassembly difficult.

To address these limitations, this study proposes a \textit{shell-type soft jig} for robotic disassembly, as illustrated in \figref{concept}. The jig combines a base jamming mechanism with an external shell structure equipped with inflatable balloons that gently press the object from all directions. This design achieves soft yet stable fixation, guiding the object toward a central upright posture and providing reliable alignment and holding performance. As a result, it reduces the need for dedicated jig design, high-accuracy perception, precise grasping, and finely tuned trajectory planning that are typically required with conventional rigid fixtures.

\begin{figure}[tb]
    \centering
    \small
    \includegraphics[width=\linewidth]{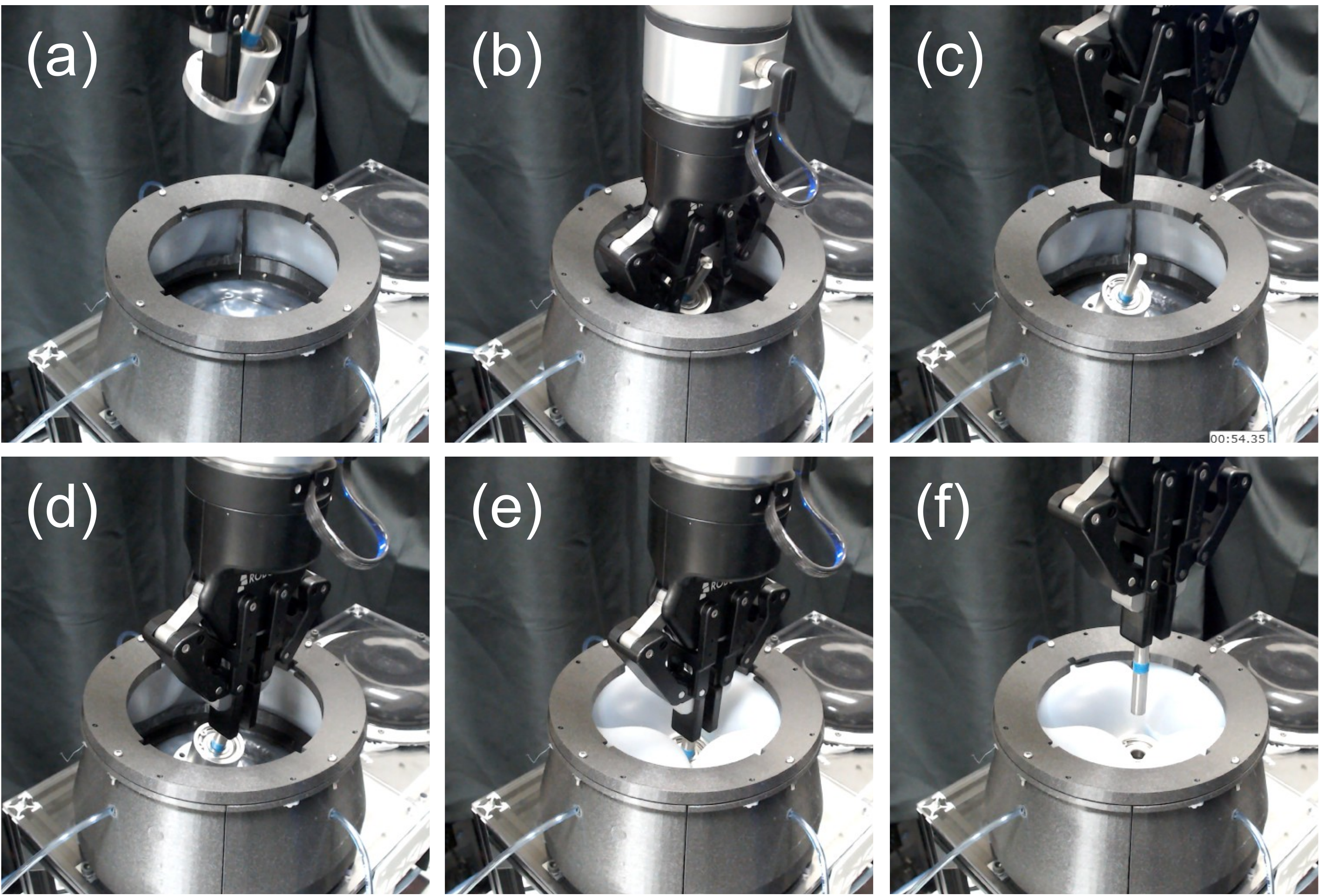}
    \caption{Assumed robotic disassembly using the proposed shell-type soft jig. 
    (a)~Transporting, (b)~Placing, (c)~Pregrasp, (d)~Grasping, (e)~Pressurization, and (f)~Pull-out.}
    \figlab{concept}
\end{figure}

\figref{concept} shows the assumed disassembly procedure using the proposed jig. 
The process consists of six steps: (a) transporting the object with the robot arm, (b) placing it onto the jig, (c) pregrasping above the object, (d) grasping the object with the gripper, (e) inflating the shell balloons to gently hold the object, and (f) performing the pull-out motion for disassembly. 
This sequence highlights how the shell-type soft jig enables both flexible fixation and robust holding during the operation.

To validate the effectiveness of the proposed jig, we conducted real-world experiments involving the disassembly of multiple representative objects. 
The experiments were designed to evaluate robustness against positional and angular misalignment, repeatability of the pull-out operation, and the holding performance of the jig. 
By applying the jig to ten representative objects, we also reveal typical success and failure cases that help identify current limitations and future research needs.

\section{Related Work}
Several jig-less operation methods~\cite{Kim2017,Naing2000} and general-purpose jig designing methods~\cite{Bi2001} have been proposed.
Grippo~\etal~\cite{Grippo1987} developed a CAD tool for generating modular fixtures from elements such as blocks and clamps.
Whybrew~\etal~\cite{Whybrew1992} proposed an automatic design system using locators, clamps, and supports.

To advance component fixturing for robotic assembly, Chan~\etal~\cite{Chan1990} proposed a perceptive reconfigurable fixture, while Hogreve~\etal~\cite{Hogreve2022} introduced an assembly wheel with robot arms for flexible fixturing.
Levi~\etal~\cite{Levi2022} later developed the \textit{SIMJig}, which uses only three actuators to control multiple clamps.
Despite their different forms, these approaches share modularity, reconfigurability, and sensory feedback.
In parallel, several researchers~\cite{pinarray3,pinarray,pinarray2} designed pin-array fixtures and grippers that conform to object shapes.
Such rigid arrays allow high-precision positioning, but their adaptability to diverse geometries is limited, reducing reusability.

To overcome this, soft robotics approaches have been explored. 
Brown~\etal~\cite{Brown2010} proposed a jamming gripper with a silicone membrane filled with granular particles. 
Following this idea, \textit{jamming jigs} were applied to robotic assembly~\cite{Kiyokawa2021jig,Sakuma2022}, where deformable membranes conform to object geometries and fix them after depressurization. 
These methods demonstrated adaptability to diverse parts but often suffered from limited holding stiffness and alignment accuracy, motivating further innovations. 
For example, Kemmotsu~\etal~\cite{Kemmotsu2024} proposed a balloon pin-array gripper with two-step adaptation, achieving stable grasping under misalignment; however, its dense pins may limit reachable space for disassembly.

Beyond pin arrays, Aoyama~\etal~\cite{Aoyama2022} developed a shell-type hybrid gripper inspired by human fingers for compact agricultural packaging, which also offers improved holding stability.
Hu~\etal~\cite{Hu2023} introduced a dual-mode enclosing gripper with tunable stiffness and high load capacity, capable of contraction- and suction-based grasping.
This paper investigates the potential of such adaptive soft jigs as fixtures for robotic disassembly.

\tabref{pros-cons} compares conventional vises/clamps, pin- or locator-based fixtures, jamming jigs, and the proposed shell-type soft jig.
Vises and clamps provide reliable fixation and alignment but lack versatility, as they must be fabricated for each geometry.
Pin- or locator-based fixtures increase versatility by repositioning supports, but fixation depends on placement resolution.
Jamming jigs enable easy fixation with air-pressure control and high shape adaptability, but their membrane has low stiffness, limiting alignment and stability.

The proposed shell-type soft jig integrates these advantages.
Deformable shells enclose the object with simple pressure control, adapt to diverse shapes, and guide it upright, ensuring accurate alignment and stable holding.
\begin{table}[t]
    \centering
    \footnotesize
        \caption{\small{Comparison of flexible fixtures and jigs for robotic manufacturing and remanufacturing}}
        \tablab{pros-cons}
        \begin{tabular}{p{7mm}p{9mm}p{9mm}p{9mm}p{9mm}} \toprule
            \multicolumn{1}{c}{Method} & \multicolumn{1}{c}{Fixing} & \multicolumn{1}{c}{Versatility} & \multicolumn{1}{c}{Alignment} & \multicolumn{1}{c}{Holding} \\ \midrule
            \multicolumn{1}{l}{Vice and clamp} & \multicolumn{1}{c}{\begin{tabular}{c}\cmark\end{tabular}} & \multicolumn{1}{c}{\begin{tabular}{c}\xmark\end{tabular}} &
            \multicolumn{1}{c}{\begin{tabular}{c}\cmark\end{tabular}} &
            \multicolumn{1}{c}{\begin{tabular}{c}\cmark\end{tabular}} \rule[-1mm]{0mm}{4mm} \\
            \multicolumn{1}{l}{Pin or locator-based} & \multicolumn{1}{c}{\begin{tabular}{c}\xmark$^{\rm *a}$\end{tabular}} & \multicolumn{1}{c}{\begin{tabular}{c}\cmark\end{tabular}} &
            \multicolumn{1}{c}{\begin{tabular}{c}\cmark\end{tabular}} &
            \multicolumn{1}{c}{\begin{tabular}{c}\cmark\end{tabular}} \rule[-1mm]{0mm}{4mm} \\
            \multicolumn{1}{l}{Jamming jig} & \multicolumn{1}{c}{\begin{tabular}{c}\cmark$^{\rm *b}$\end{tabular}} & \multicolumn{1}{c}{\begin{tabular}{c}\cmark$^{\rm *c}$\end{tabular}} & 
            \multicolumn{1}{c}{\begin{tabular}{c}\xmark$^{\rm *d}$\end{tabular}} & 
            \multicolumn{1}{c}{\begin{tabular}{c}\xmark\end{tabular}} \rule[-1mm]{0mm}{4mm} \\ 
            \multicolumn{1}{l}{Shell-type soft jig} & \multicolumn{1}{c}{\begin{tabular}{c}\cmark$^{\rm *e}$\end{tabular}} & \multicolumn{1}{c}{\begin{tabular}{c}\cmark$^{\rm *f}$\end{tabular}} & \multicolumn{1}{c}{\begin{tabular}{c}\cmark$^{\rm *g}$\end{tabular}} & \multicolumn{1}{c}{\begin{tabular}{c}\cmark\end{tabular}} \rule[-1mm]{0mm}{4mm} \\ 
            \bottomrule
        \end{tabular}
        \begin{tablenotes}
          \item[a]\footnotesize{$^{\rm *a}$ Fixation versatility depends on pin or locator placement resolution}
          \item[b]\footnotesize{$^{\rm *b}$ On-off control of air pressure}
          \item[c]\footnotesize{$^{\rm *c}$ Deformable body fits the object shape}
          \item[d]\footnotesize{$^{\rm *d}$ Stiffness of the malleable membrane is lower than the rigid jig surface}
          \item[e]\footnotesize{$^{\rm *e}$ On-off control of air pressure}
          \item[f]\footnotesize{$^{\rm *f}$ Surrounding deformable bodies fit the object shape}
          \item[g]\footnotesize{$^{\rm *g}$ Automatically guided upright at the center when pressed}
        \end{tablenotes}
\end{table}

\section{Problem Setting}
\figref{concept} shows an example disassembly task using the proposed shell-type soft jig, specifically, extracting a shaft from a bearing. In this study, we assume that a disassembly motion plan is given in advance. The plan specifies the grasping configuration with a two-finger gripper and the corresponding trajectory of a single-arm manipulator equipped with the gripper.

The procedure for using the shell-type soft jig follows six steps. First, according to the planned trajectory, the object to be fixed is grasped by the two-finger gripper and transported to the jig (\figref{concept}~(a)). The robot then places the object onto the bottom-mounted jamming jig, which is depressurized to achieve initial fixation (\figref{concept}~(b)). After the object is positioned, the gripper is released and moved away, returning to its pre-grasp state (\figref{concept}~(c)). Next, the gripper re-approaches to grasp only the component to be removed (\figref{concept}~(d)). 

To softly hold the fixed object during disassembly, the air chamber embedded inside the shell-shaped outer wall is pressurized, inflating the silicone membranes to apply gentle surrounding pressure (\figref{concept}~(e)). This balloon-based holding mechanism aligns the object toward the center and stabilizes its posture. Finally, while the jig maintains the object in place, the target component is extracted by the gripper following the motion plan (\figref{concept}~(f)).

In this study, the problem setting assumes a single-arm manipulator with a two-finger gripper, a pre-planned disassembly trajectory, and a task focused on pull-out operations. The role of the shell-type soft jig is to provide robust alignment and holding of the fixed object, thereby reducing the dependence on precise perception, grasping accuracy, and finely tuned trajectory planning.

\section{Proposal}
\subsection{Structure of the Shell-Type Soft Jig}
\begin{figure}[tb]
    \centering
    \small
    \begin{minipage}[tb]{0.62\linewidth}
      \centering
        \includegraphics[keepaspectratio, width=\linewidth]{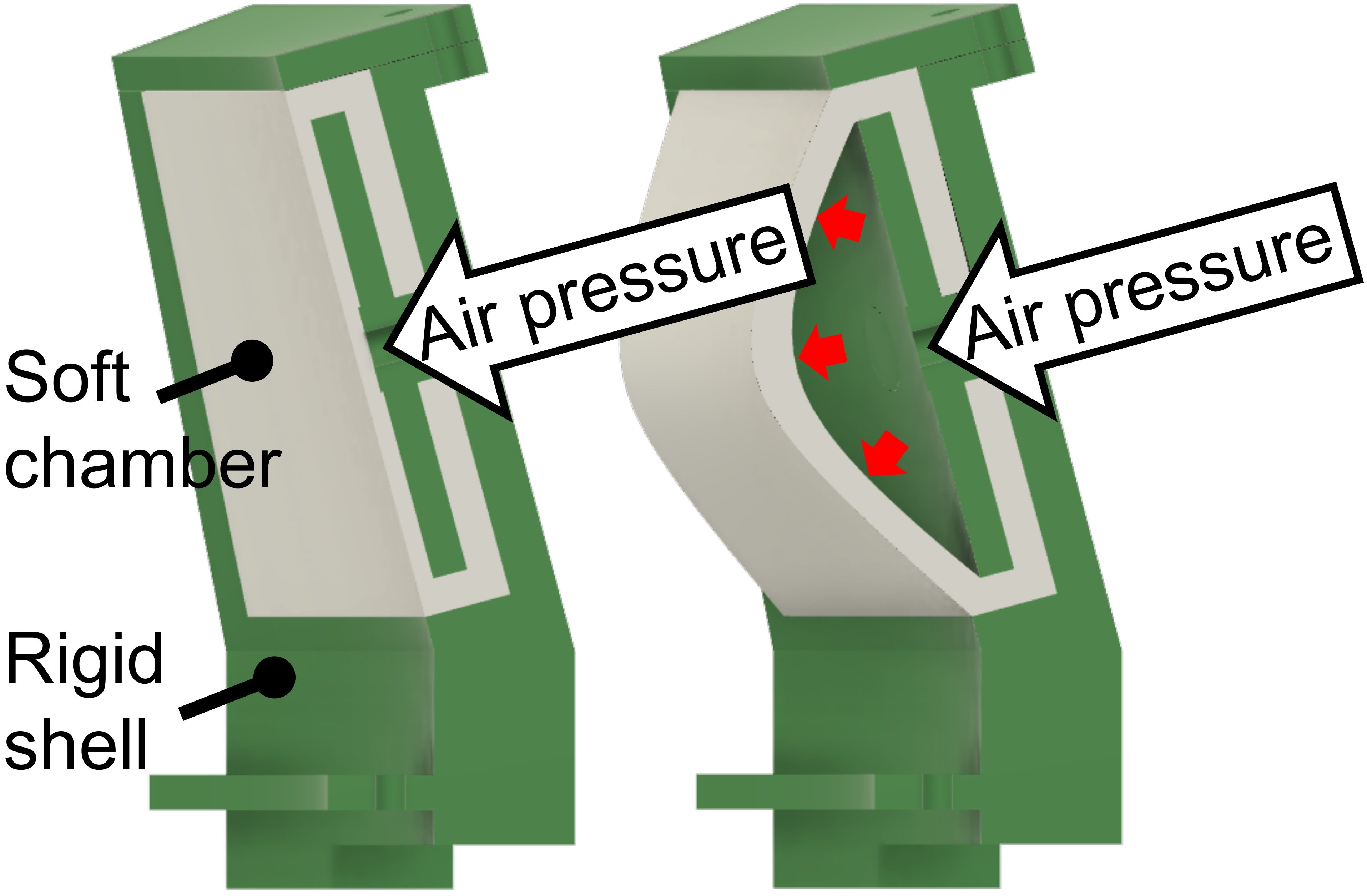}
        \subcaption{Pressurization mechanism}
    \end{minipage}
    \begin{minipage}[tb]{0.36\linewidth}
        \centering
        \includegraphics[keepaspectratio, width=\linewidth]{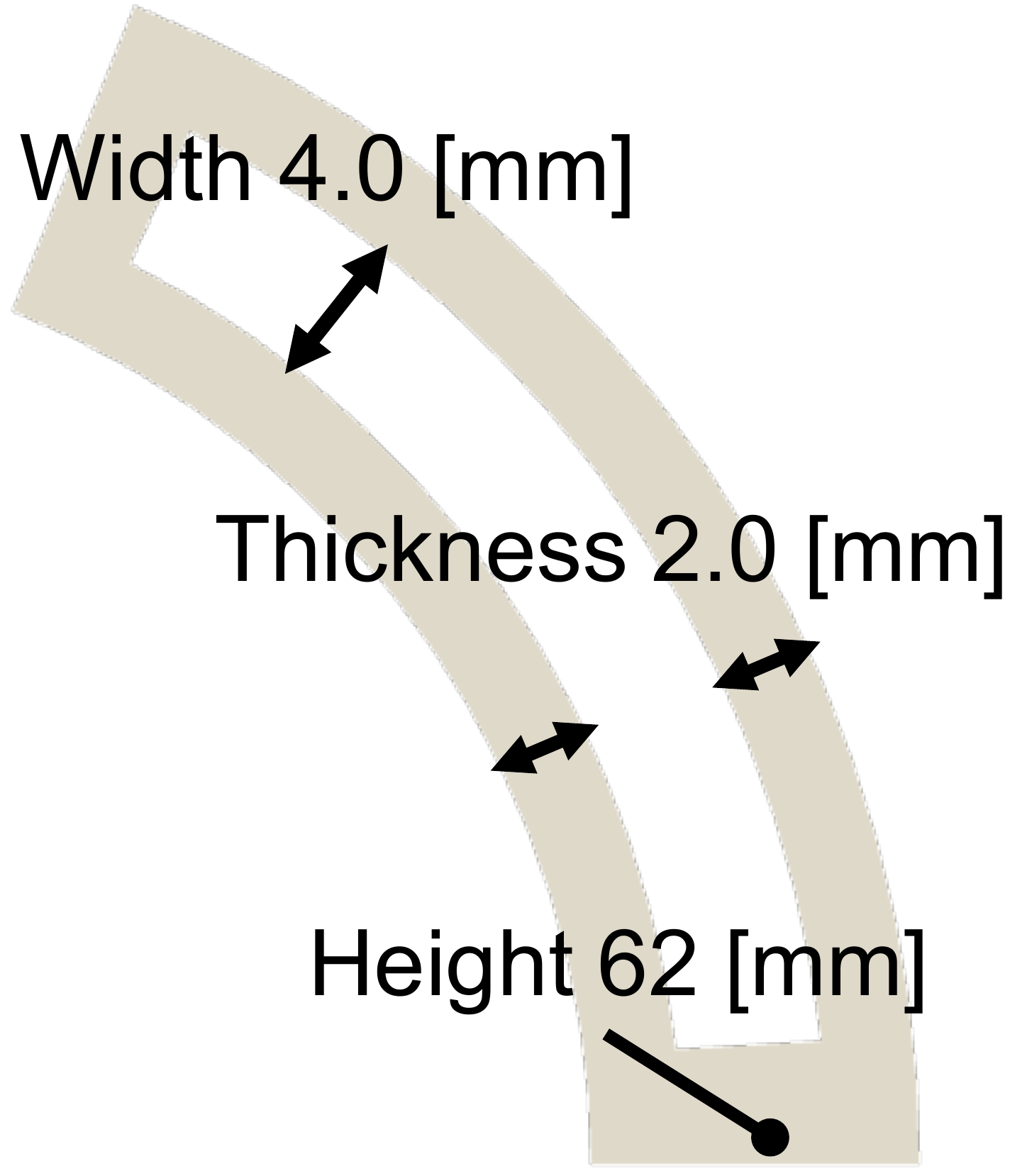}
        \subcaption{Design parameters}
    \end{minipage}
    \caption{Function and design of the air chamber}
    \figlab{chamber}
\end{figure}

\figref{chamber}(a) shows the overall structure of the proposed shell-type soft jig. 
The green region represents the rigid outer shell, which surrounds the object to be fixed and held. 
The white region denotes an air chamber covered with a silicone membrane, which can be pressurized or depressurized. 
When air pressure is applied through the connected inlet, the flexible membrane inflates toward the object. 
By using an external compressor, an electro-pneumatic proportional valve, and a control unit, the internal pressure of each chamber can be precisely regulated. 

\figref{chamber}(b) presents the design parameters of the chamber. 
The dimensions of the chamber thickness, width, and height define the amount of membrane expansion and the resulting contact area with the object. 
These design parameters are critical because the stiffness and maximum deformation of the membrane depend on its geometry and material properties. 
Selecting appropriate dimensions ensures that the membrane can expand sufficiently to generate stable holding forces without excessive interference or loss of compliance.

Multiple such modules can be arranged around the object, enabling fixation from multiple directions. 
Increasing the number of modules improves holding performance, as the membranes conform to complex surface geometries by enveloping the object from several sides when inflated. 
However, too many modules complicate pneumatic wiring and reduce the expansion of each membrane due to spatial constraints. 
Therefore, careful design is required to balance the number of modules with the system's control complexity and holding capability.

\subsection{Configuration for Caging-Based Object Holding}
\begin{figure}[tb]
    \centering
    \small
    \begin{minipage}[tb]{0.49\linewidth}
        \centering
        \includegraphics[keepaspectratio, width=\linewidth]{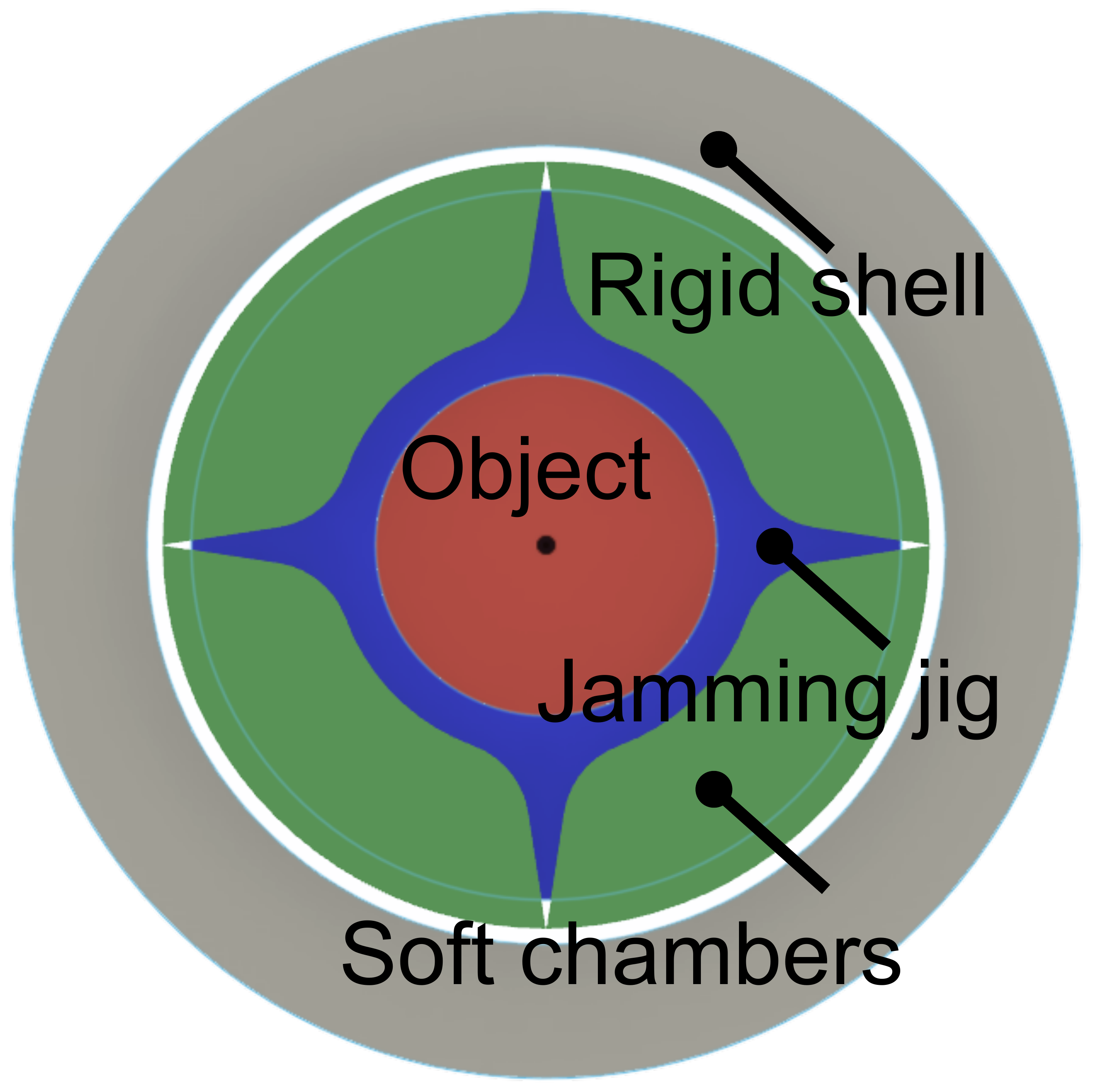}
        \subcaption{Caging concept}
    \end{minipage}
    \begin{minipage}[tb]{0.49\linewidth}
        \centering
        \includegraphics[keepaspectratio, width=\linewidth]{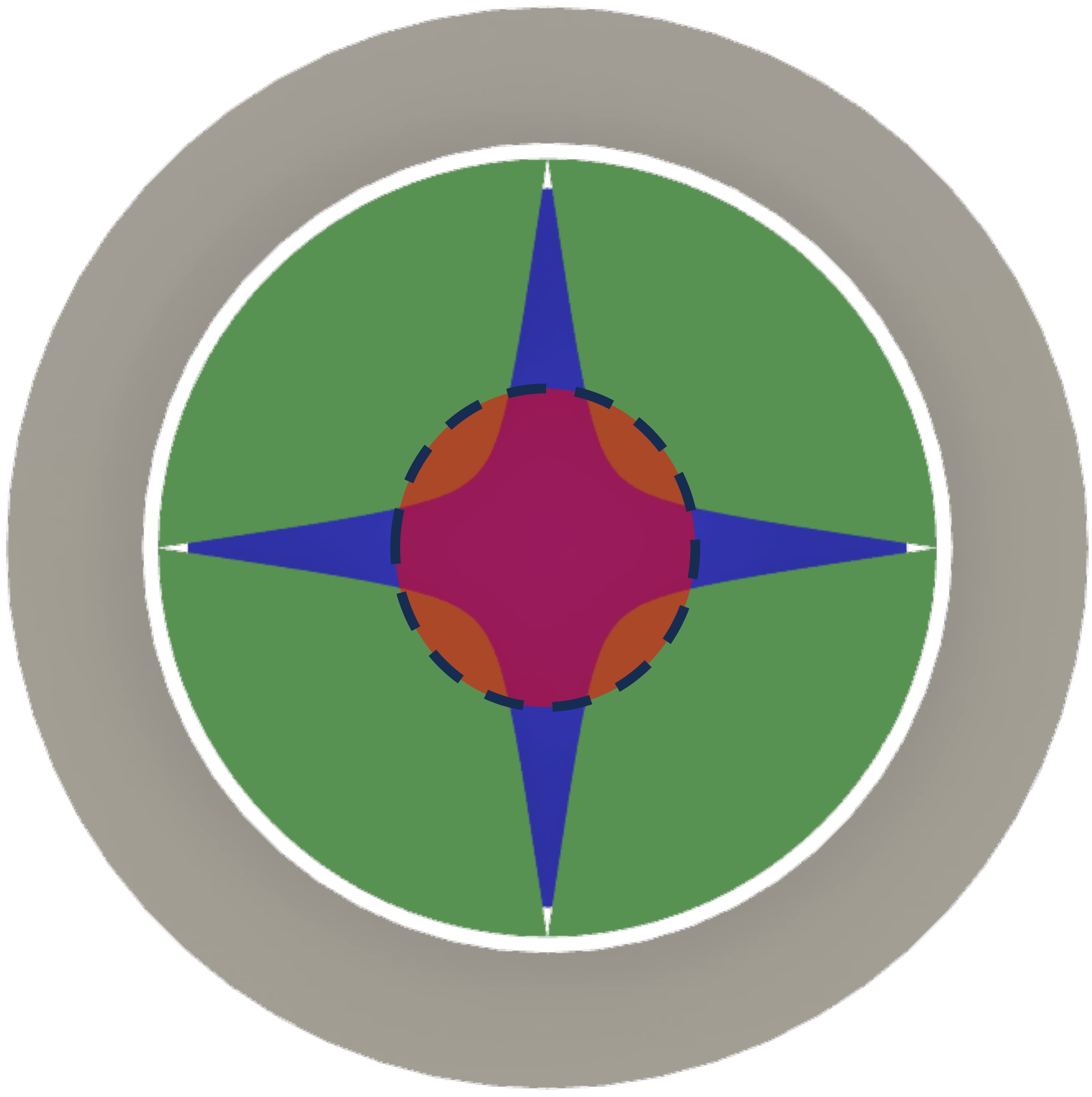}
        \subcaption{Rigid vs. soft shells}
    \end{minipage}
    \caption{Caging-based holding principle}
    \figlab{caging}
\end{figure}

In disassembly tasks, softly fixing the target object can help absorb pose errors, allowing more robust operations. 
The proposed method employs object holding based on caging in a two-dimensional plane. \figref{caging}(a) illustrates it.
This concept assumes that the object can be placed inside a rigid boundary without immediate contact, and then stabilized by the deformation of surrounding soft chambers. 

\figref{caging}(b) contrasts the case of rigid shells only with that of shells equipped with soft chambers. 
In the rigid-only case, the object can occupy a free configuration without contacting the boundary, meaning caging is not guaranteed. 
By contrast, when soft chambers are integrated, the membranes expand and close the free space, ensuring that the object remains enclosed and aligned even under small pose variations. 
This comparison highlights the importance of combining rigid guidance with soft compliance to realize robust caging-based holding.

Maeda~\etal~\cite{caging} defined the caging conditions for object grasping, particularly when contact interfaces involve deformable materials. 
They provide the theoretical basis for stable holding with soft contact surfaces:
\begin{align}
    C_\mathrm{free-ICS} \neq \emptyset,  \forlab{rigid} \\   
    q_\mathrm{obj} \in C_\mathrm{free-ICS},  \forlab{caging} \\
    C^{'}_\mathrm{free-ICS} = \emptyset.  \forlab{soft}
\end{align}
Here, $q_\mathrm{obj}$ is the object configuration, $C_\mathrm{free\text{-}ICS}$ the space without rigid contact, and $C^{'}_\mathrm{free\text{-}ICS}$ the space without soft contact. 
Conditions \forref{rigid} and \forref{caging} (\figref{caging}(a)) mean the object can be placed without touching the rigid shell, while \forref{soft} (\figref{caging}(b)) implies the object is always enclosed by undeformed membranes. 

Thus, the hardware must satisfy all three conditions to enable caging-based holding. 
In practice, the shell diameter should permit initial placement without rigid contact. The membrane must then expand sufficiently, depending on its material, thickness, and height, to fill the inner space upon pressurization.

\subsection{Fabrication Process}
\begin{figure}[tb]
    \centering
    \small
    \begin{minipage}[tb]{0.49\linewidth}
      \centering
        \includegraphics[keepaspectratio, width=\linewidth]{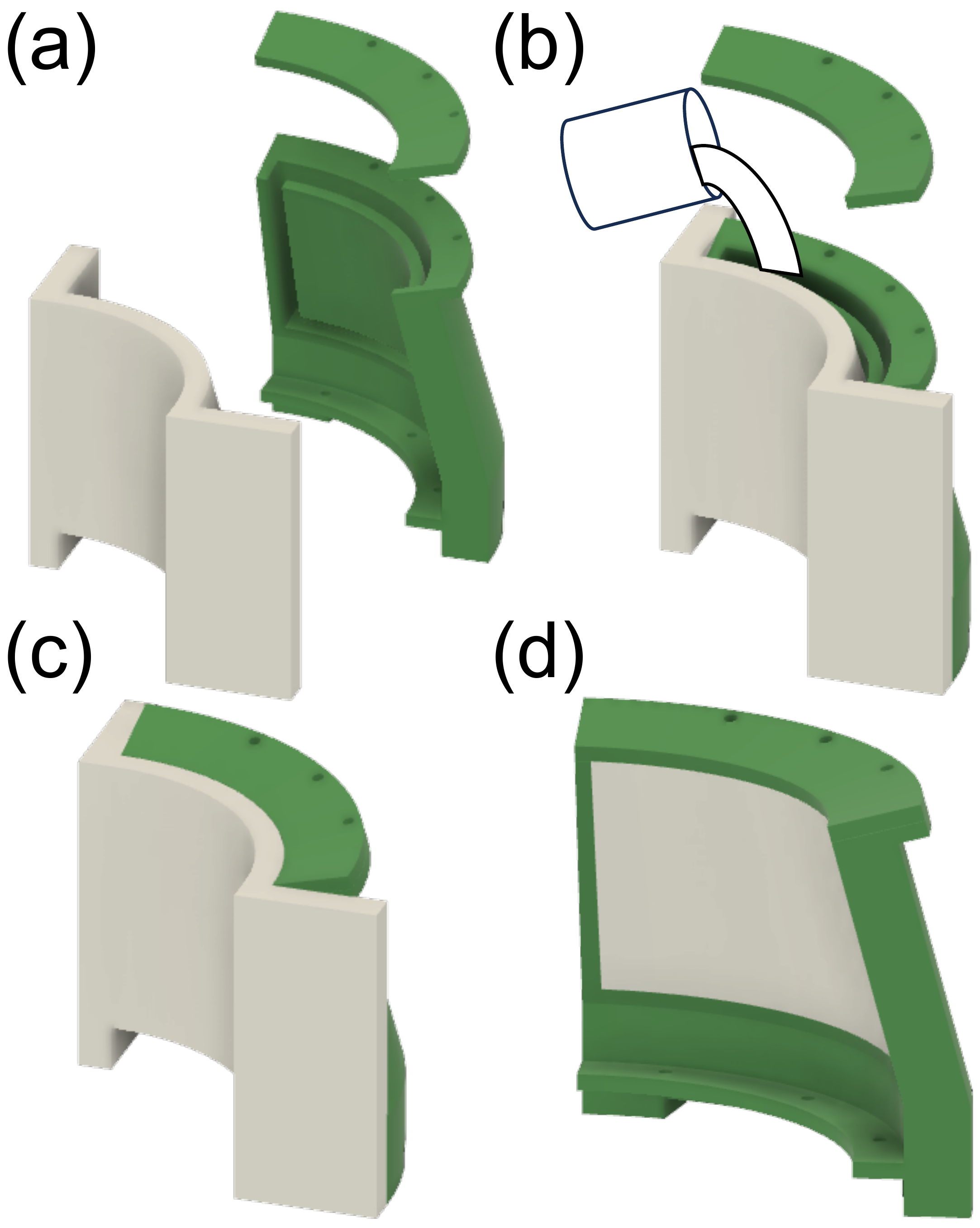}
        \subcaption{Module molding}
    \end{minipage}
    \begin{minipage}[tb]{0.45\linewidth}
        \centering
        \includegraphics[keepaspectratio, width=\linewidth]{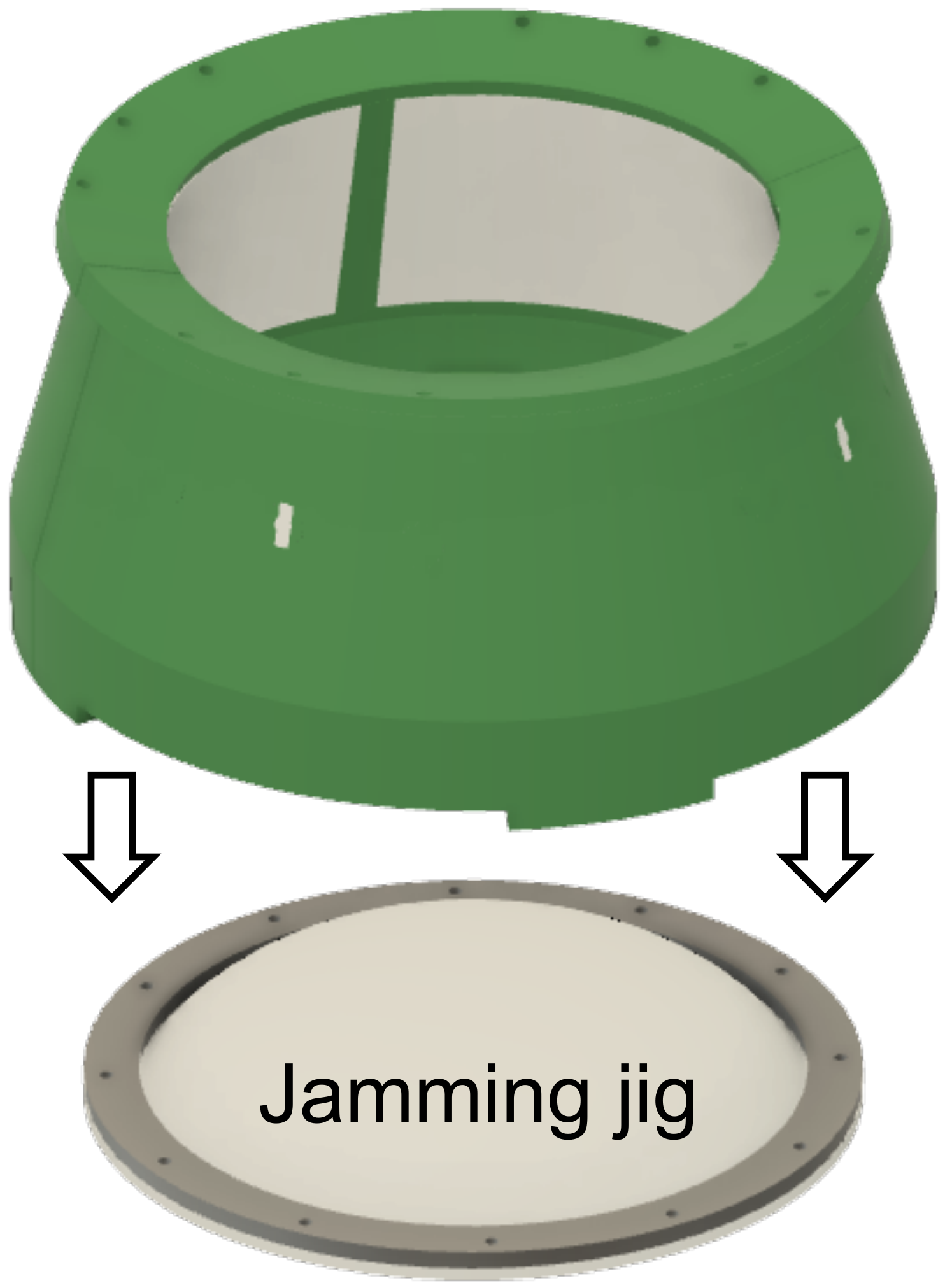}
        \subcaption{Assembly}
    \end{minipage}
    \caption{Fabrication process of shell-type soft jig}
    \figlab{fabrication}
\end{figure}

\figref{fabrication} draws the fabrication process of the proposed jig. 
First, the rigid shell and the soft chamber parts are manufactured separately, for example through molding with silicone membranes and resin structures (\figref{fabrication}(a)). 

Next, the soft chambers are integrated into the rigid shell, and pneumatic inlets are embedded for pressurization control. 
Finally, the completed shell module is combined with the bottom jamming jig to form the full jig assembly (\figref{fabrication}(b)). 
This modular fabrication approach simplifies replacement and scaling: the number of soft chamber units can be adjusted depending on the object size, and damaged modules can be exchanged individually without reconstructing the entire jig.

\begin{figure}[tb]
    \centering
    \small
    \includegraphics[keepaspectratio, width=\linewidth]{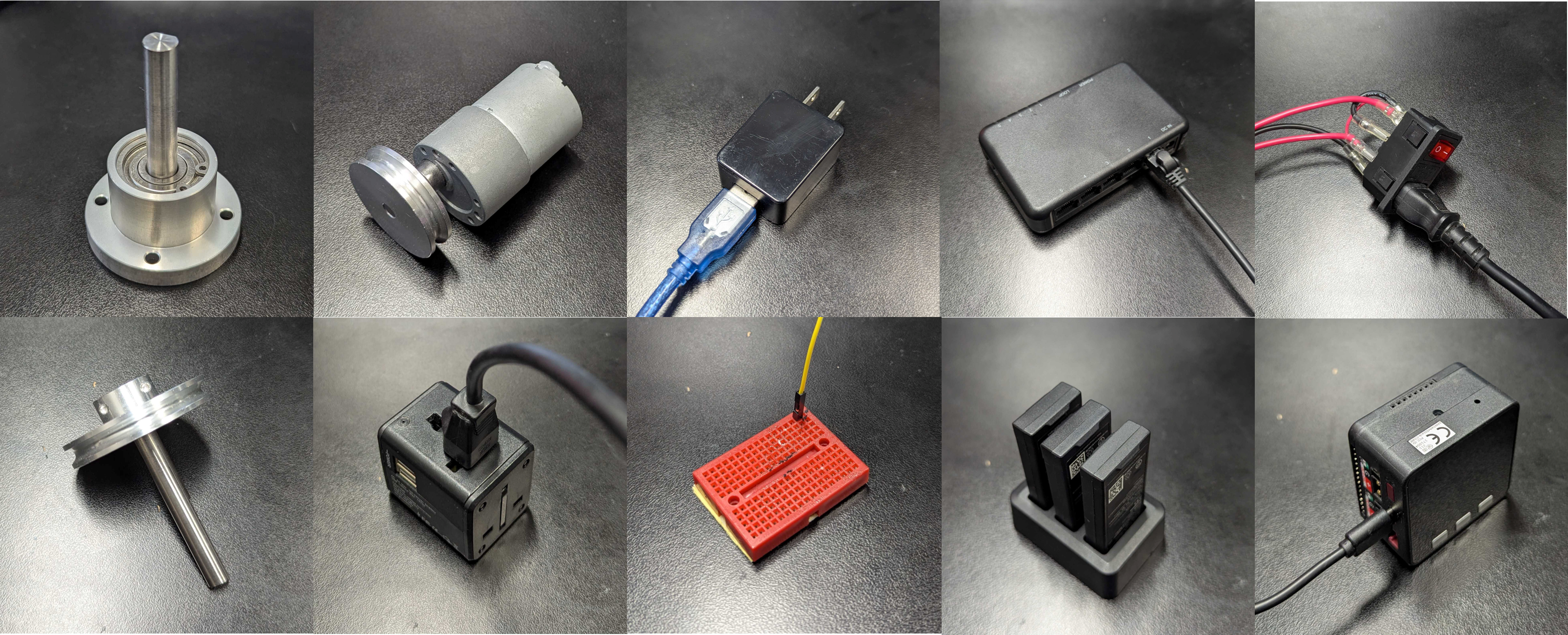}
    \caption{Experimental objects, including shaft--bearing, motor--pulley, USB--adapter, LAN--hub, AC--switch, pulley--shaft, AC--adapter, wire--board, battery--charger, and USB--computer.}
    \figlab{disassembly}
\end{figure}
\begin{figure}[tb]
    \centering
    \small
    \includegraphics[width=\linewidth]{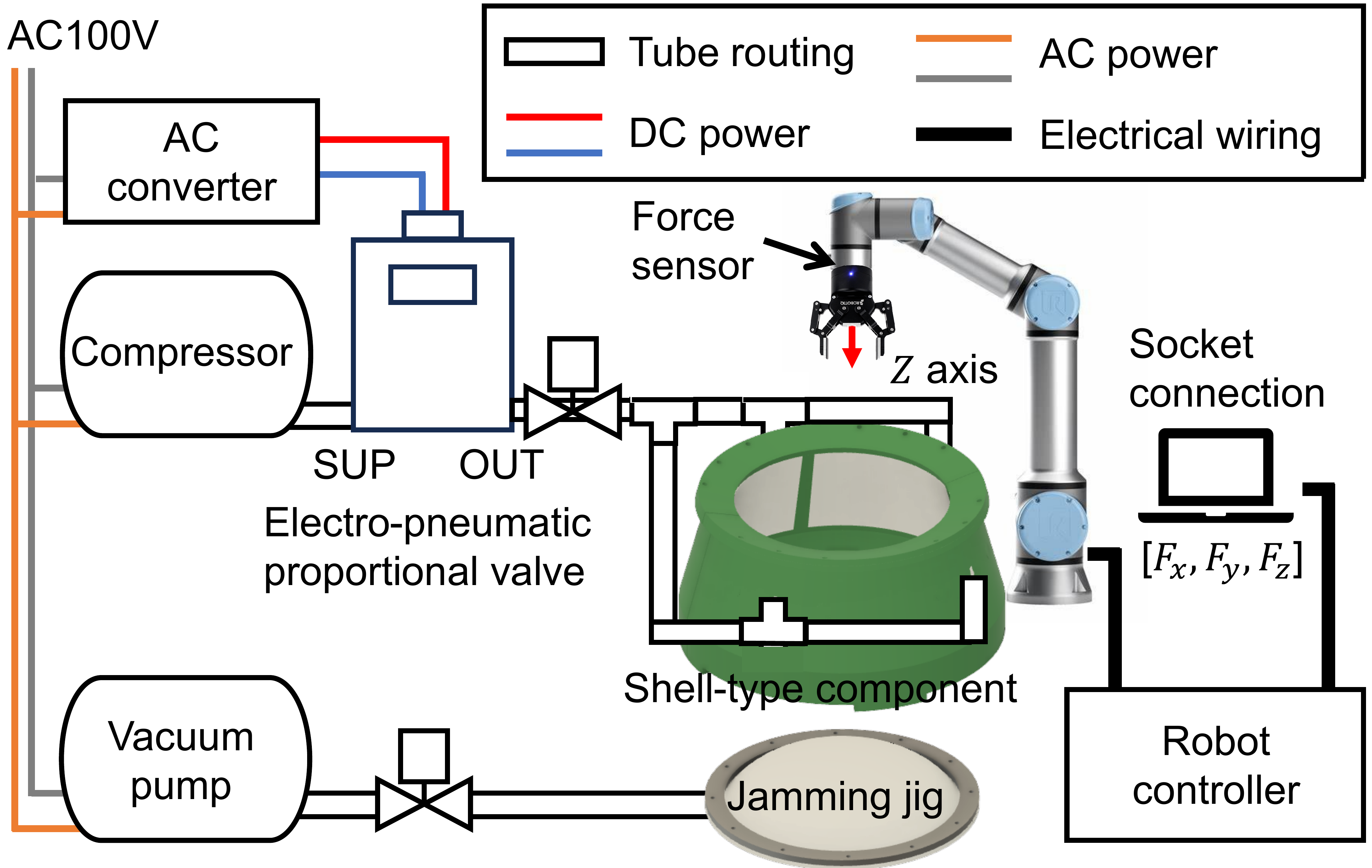}
    \caption{Overview of the implemented system}
    \figlab{system}
\end{figure}
\begin{figure}[tb]
    \centering
    \small
    \begin{minipage}[tb]{0.32\linewidth}
        \centering
        \includegraphics[keepaspectratio, width=\linewidth]{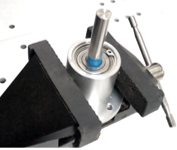}
        \subcaption{Vise}
    \end{minipage}
    \begin{minipage}[tb]{0.32\linewidth}
        \centering
        \includegraphics[keepaspectratio, width=\linewidth]{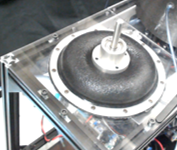}
        \subcaption{Jamming jig}
    \end{minipage}
    \begin{minipage}[tb]{0.32\linewidth}
        \centering
        \includegraphics[keepaspectratio, width=\linewidth]{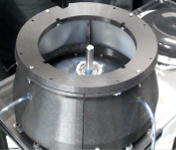}
        \subcaption{Proposed}
    \end{minipage}
    \caption{Comparative object fixing methods}
    \figlab{exp-condition}
\end{figure}

\section{Evaluating Holding Performance}
\subsection{Setup}
To evaluate the robustness and holding performance of the proposed shell-type soft jig, we conducted disassembly experiments using two robot platforms depending on the task.
\figref{disassembly} shows the experimental targets consisting of ten assembled objects: a shaft–bearing, a motor–pulley, a USB–adapter, a LAN–hub, an AC–switch, a pulley–shaft, an AC–adapter, a wire–board connection, a battery–charger, and a USB–computer connector.
These objects were chosen to cover a variety of geometries and connection types relevant to practical disassembly scenarios.

For robustness evaluation, we used a LBR iiwa 14 R820 (KUKA) robot arm equipped with a Hand-E (Robotiq) gripper. 
This lightweight, torque-controlled robot enabled safe and consistent execution of multiple trials. 
For holding performance evaluation by measuring force applications, we employed a UR5e (Universal Robots) robot arm with an integrated 3-axis wrist force/torque sensor, combined with a 2F-85 (Robotiq) gripper. 
Force data were transmitted via socket communication between the UR5e controller and an external PC for synchronized recording and analysis.

The proposed shell-type jig was implemented as a rigid frame enclosing air chambers. 
The chamber pressures were regulated using electro-pneumatic proportional valves connected to a DC power supply and an air compressor. 
A jamming-based jig was also prepared, consisting of a flexible membrane connected to a vacuum pump for depressurization-based fixation. 
A conventional vise served as a baseline rigid fixture.

Based on the planned disassembly experiments and the size constraints of both the target object and the shell-type soft jig, the number of modules was limited to four. 
This corresponds to the maximum that can simultaneously contact the object’s surface, ensuring sufficient membrane contact while keeping the experimental setup feasible. 
To satisfy the condition \forref{soft}, we used Dragon Skin 10, a high-performance silicone rubber manufactured by Smooth-On Inc., known for its relatively high tensile strength and 10\% modulus of elasticity\footnote{\url{https://www.sac-corp.co.jp/products/DRAGON_SKIN_TI.pdf}}. 
\figref{chamber}(b) shows the dimensions of the outer shell used to form the air chamber were determined through trial-and-error preliminary experiments. 
Specifically, the width of the inner cavity of the shell was set to 4.0\,mm, the wall thickness to 2.0\,mm, and the height to 62\,mm.

\figref{system} shows an overview of the entire experimental system. 
It illustrates how the compressor and vacuum pump are connected to the electro-pneumatic proportional valves for regulating chamber pressure, how the robot controller and PC communicate via socket connection to record force data, and how the shell-type soft jig and the jamming jig are integrated into the system. 
This diagram summarizes the power supply, pneumatic routing, sensing, and control flow of the implemented setup.

\subsection{Procedure}
In each trial, the target component was first grasped by the gripper and then fixed using one of three jigs, as illustrated in \figref{exp-condition}: (a) vise-based jig, (b) jamming-based soft jig, and (c) proposed shell-type soft jig.

For robustness evaluation against initial pose errors, pull-out operations were performed with angular deviations between the robot’s pulling trajectory and the ideal extraction direction of the target. The angular deviation was incremented in $5^{\circ}$ steps from $0^{\circ}$ to $25^{\circ}$, and the success or failure of each attempt was recorded. From these results, the maximum allowable angular deviation was identified for each jig.

For holding performance evaluation, we conducted 10 trials for three representative objects, the shaft–bearing, motor–pulley, and USB–adapter, under angular deviations of $0^{\circ}$, $5^{\circ}$, and $10^{\circ}$. 

\subsection{Allowable Angular Deviation and Robustness}
\begin{table}[tb]
    \centering
    \footnotesize
    \caption{\small{Successful conditions of the pull-out different objects}}
    \tablab{pull_out_result}
    \begin{tabular}{p{18mm}p{15mm}ccccccr} \toprule
        \multicolumn{1}{c}{Target} & \multicolumn{1}{c}{Method} & \multicolumn{6}{c}{Angular difference [deg]} \\ \cmidrule(lr){3-8}
        & & 0 & 5 & 10 & 15 & 20 & 25 \\ \midrule
        \multirow{3}{*}{shaft--bearing} 
            & Vise & \cmark & \cmark &  &  &  &  \\
            & Jamming jig & \cmark & \cmark & \cmark &  &  &  \\
            & Proposed & \cmark & \cmark & \cmark & \cmark & \cmark & \cmark \\  \midrule
        \multirow{3}{*}{motor--pulley} 
            & Vise & \cmark & \cmark &  &  &  &  \\
            & Jamming jig & \cmark & \cmark & \cmark &  &  &  \\
            & Proposed & \cmark & \cmark & \cmark & \cmark & \cmark & \cmark \\ \midrule
        \multirow{3}{*}{USB--adapter} 
            & Vise & \cmark & \cmark &  &  &  &  \\
            & Jamming jig & \cmark & \cmark &  &  &  &  \\
            & Proposed & \cmark & \cmark & \cmark & \cmark & \cmark & \cmark \\ \midrule
        \multirow{3}{*}{LAN--hub} 
            & Vise & \cmark &  &  &  &  &  \\
            & Jamming jig & \cmark &  &  &  &  &  \\
            & Proposed & \cmark & \cmark & \cmark &  &  &  \\ \midrule
        \multirow{3}{*}{AC--switch} 
            & Vise &  &  &  &  &  &  \\
            & Jamming jig &  &  &  &  &  &  \\
            & Proposed &  &  &  &  &  &  \\ \bottomrule
        \multirow{3}{*}{pulley--shaft} 
            & Vise & \cmark & \cmark &  &  &  &  \\
            & Jamming jig &  &  &  &  &  &  \\
            & Proposed &  &  &  &  &  & \\  \midrule
        \multirow{3}{*}{AC--adapter} 
            & Vise & \cmark & \cmark &  &  &  &  \\
            & Jamming jig & \cmark & \cmark & \cmark &  &  &  \\
            & Proposed & \cmark & \cmark & \cmark & \cmark & \cmark & \cmark \\ \bottomrule
        \multirow{3}{*}{wire--board} 
            & Vise & \cmark &  &  &  &  &  \\
            & Jamming jig & \cmark &  &  &  &  &  \\
            & Proposed & \cmark & \cmark & \cmark &  &  &  \\ \bottomrule
        \multirow{3}{*}{battery--charger} 
            & Vise & \cmark & \cmark &  &  &  &  \\
            & Jamming jig & \cmark & \cmark & \cmark &  &  &  \\
            & Proposed & \cmark & \cmark & \cmark & \cmark & \cmark & \cmark  \\ \midrule
        \multirow{3}{*}{USB--computer} 
            & Vise & \cmark & \cmark &  &  &  &  \\
            & Jamming jig & \cmark & \cmark & \cmark &  &  &  \\
            & Proposed & \cmark & \cmark & \cmark & \cmark & \cmark & \cmark  \\ \midrule
    \end{tabular}
    \begin{tablenotes}
      \footnotesize
      \item[\cmark] \cmark~represents successful condition.
    \end{tablenotes}
\end{table}

\tabref{pull_out_result} shows the successful conditions of pull-out operations for the ten objects. A rigid fixture such as a vise, which completely constrains the object’s pose, enabled extraction only under strict alignment. For example, the shaft--bearing and motor--pulley could be pulled out at $0^{\circ}$ but often failed even at larger angular deviations after $10^{\circ}$, as the target object slipped from the gripper during pulling. Similar limitations were observed for the USB--adapter, LAN--hub, pulley--shaft, AC--adapter, wire--board, battery--charger, and USB--computer, with success restricted to nearly perfect alignment.

In contrast, the jamming-based soft jig expanded the tolerance range by exploiting its deformable membrane. For the shaft–-bearing, motor--pulley, AC--adapter, battery--charger, and USB--computer, successful extraction was achieved up to $10^{\circ}$. However, the relatively low stiffness of the membrane limited holding stability, and extraction failed when the angular deviation became larger.

The proposed shell-type soft jig achieved the broadest tolerance to misalignment.
For the shaft--bearing, motor--pulley, USB--adapter, AC--adapter, battery--charger, and USB--computer, successful extractions were consistently achieved up to $25^{\circ}$, accompanied by stable guidance toward the upright central position during pulling.
For the LAN--hub and wire--board, extractions remained successful up to $10^{\circ}$, although performance degraded at larger angles.
In contrast, no method succeeded for the AC--switch and pulley--shaft under the tested conditions, indicating the need for further design improvements or integration with planning strategies.

Overall, these results indicate that the proposed jig provides the highest robustness against angular misalignment among the three methods, demonstrating its effectiveness in improving the success rate of disassembly tasks under non-ideal conditions. Please refer to the video for successful examples of pull-out operations for representative objects and the failure case for the AC--switch and pulley--shaft.

\subsection{Force Measurement Results and Holding Performance}
\begin{figure}[tb]
    \centering
    \small
    \begin{minipage}[tb]{0.48\linewidth}
        \centering
        \includegraphics[keepaspectratio, width=\linewidth]{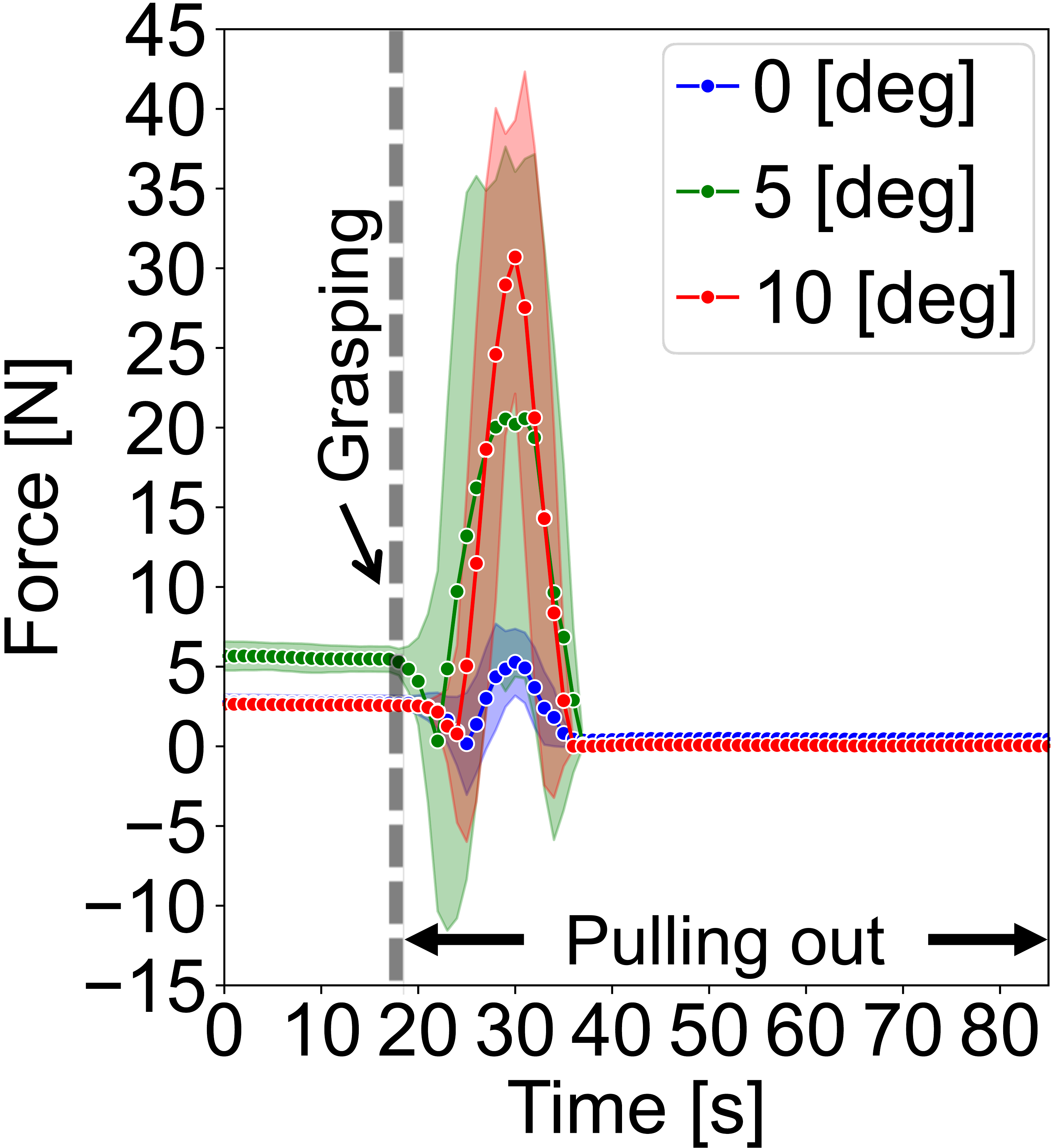}
        \subcaption{Vise}
    \end{minipage}
    \begin{minipage}[tb]{0.48\linewidth}
        \centering
        \includegraphics[keepaspectratio, width=\linewidth]{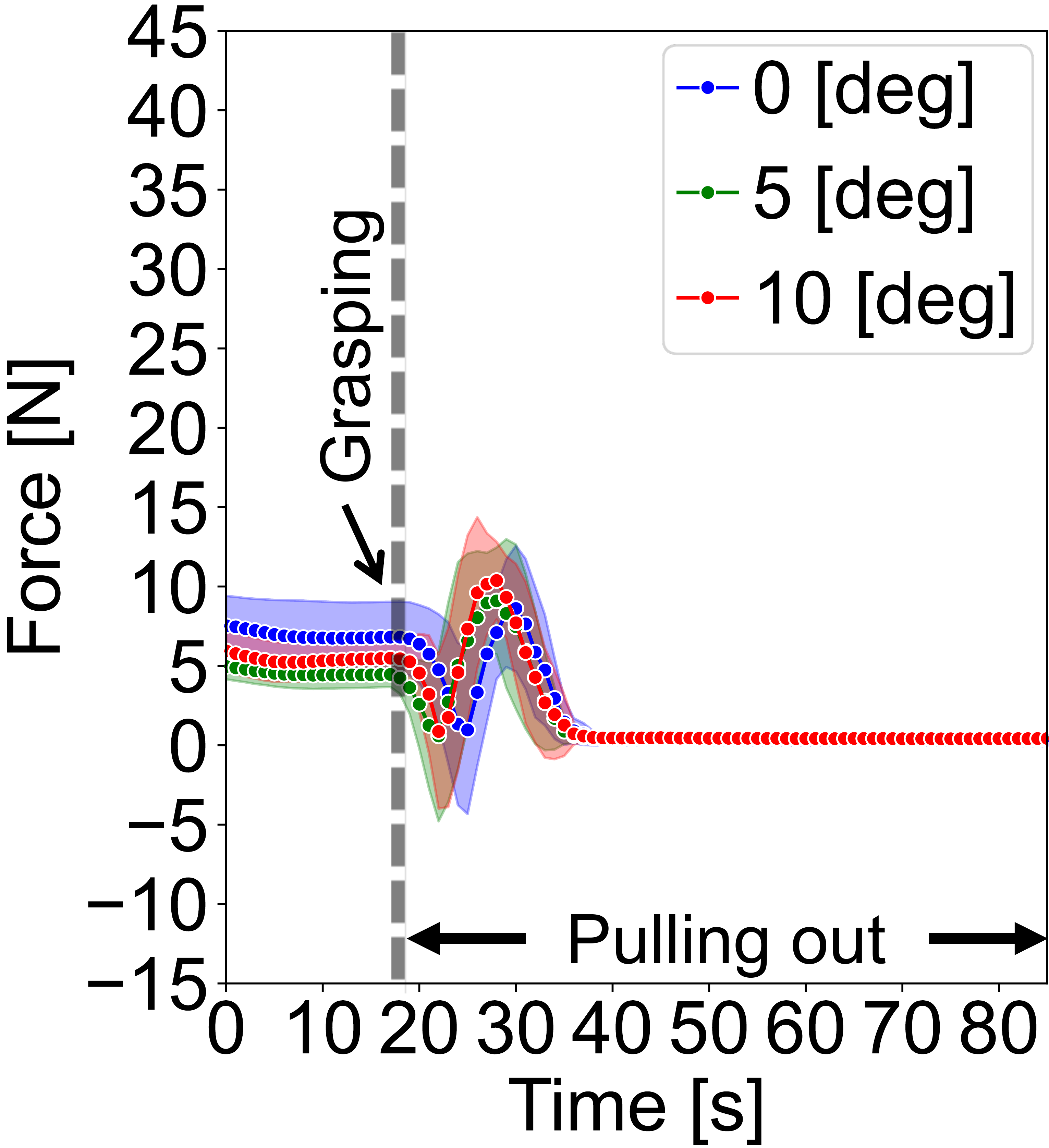}
        \subcaption{Proposed}
    \end{minipage}
    \caption{Force values measured at the arm's wrist during pull-out operations}
    \figlab{force}
\end{figure}
\figref{force} shows the results of force measurement. The figure shows the absolute values of the vertical force measured at the robot wrist during the pull-out motion. For each angular deviation condition, the mean and standard deviation over 10 trials are plotted as solid lines and shaded regions, respectively.
\figref{force}~(a) and \figref{force}~(b) show the results for the vise-based method and the proposed method, respectively.

In the case of the vise-based method, even with a 5-degree angular deviation, the force applied to the arm sharply increased at the moment the target object was pulled against the fixed component and then quickly returned to near 0~[N]. This behavior indicates that the object slipped from the gripper and that extraction failed. This is likely due to the rigid constraint of the vise, which prevented any positional adjustment of the object during the operation.

In contrast, with the proposed method, approximately 10~[N] of force was applied during extraction, but no excessive force was observed, and all 10 trials resulted in successful extractions. The standard deviation across the trials was relatively small, indicating high repeatability of the disassembly task.
This outcome suggests that the soft membrane of the air chamber conformed to the shape of the fixed component and passively adapted during deformation. As the fixed object's pose was allowed to change slightly, the pose of the target object also adjusted along the pulling trajectory, resulting in successful extractions.

\section{Discussion}
Flexible or soft fixtures have been studied as alternatives to rigid ones, but they often reduce positional accuracy and make pose estimation difficult~\cite{Sakuma2022}. The proposed shell-type soft jig alleviates this by pressing the object uniformly with balloon chambers, which guides it toward an upright posture and reduces the need for precise localization.

Experiments on ten objects of varied geometry confirmed that the jig compensates for angular and positional errors, improving success rates compared to vise- and jamming-based fixtures. Validation with different robot platforms and grippers further suggests its general applicability. Nevertheless, several limitations were observed. For the LAN--hub, extraction failed when the gripper missed the tab, highlighting that connectors with mechanical locks still demand accurate approach. For the AC--switch, no success was achieved because the required holding force exceeded the jig’s capacity. Additional difficulties appeared in the pulley--shaft, where the large pulley prevented balloon contact with the shaft and the high center of gravity caused tipping before inflation. The wire--board also proved problematic: slight tilting bent the wire inside the socket, preventing removal. These cases show that connectors with locks or deformable leads remain challenging for balloon-based holding.

Scalability and durability are other important issues. While the tested parts were within hand-sized ranges, larger or smaller objects will require adaptations such as modular balloon units, pressure regulation, or different surface materials.
For the durability, we might be able to refer to previous work~\cite{Tadakuma2020}.
Such strategies could broaden applicability to a wider set of components. 

Overall, the jig reduces dependence on high-precision jig design, perception, and trajectory planning, but hardware and control improvements are still necessary for rigid or tightly constrained connectors. Future work will focus on enhancing balloon design, modularity, and control strategies to extend the range of objects that can be reliably handled.

\section{Conclusion}
We proposed a shell-type soft jig that integrates a jamming base with balloon side chambers to hold diverse objects for robotic disassembly. 
Experiments on ten objects showed higher success rates than vise- or jamming-based fixtures, as the balloon mechanism improved robustness to pose errors and provided stable holding during pulling. 
This reduces the need for precise jig design, accurate localization, and strict trajectory planning.

The evaluation across ten diverse objects highlighted both successful and failed disassembly cases, pointing out limitations while also demonstrating the jig’s potential for broader application.
Limitations were observed for connectors with locking tabs (LAN--hub) and for cases requiring greater holding force (AC--switch). 
Future work includes improving balloon design, durability, friction materials, and modular scalability to handle different object sizes. 
These advances will broaden the jig’s applicability for versatile and reliable robotic disassembly.

\section*{Acknoledgement}
This work was supported by the New Energy and Industrial Technology Development Organization (NEDO) project JPNP23002. 

\bibliographystyle{IEEEtran}
\footnotesize
\bibliography{reference}

\end{document}